\documentclass[10pt,twocolumn,letterpaper]{article}
\usepackage[pagenumbers]{cvpr}      % Camera-ready: named authors and no line numbers.
\usepackage[T1]{fontenc}
\usepackage{amsmath,amssymb,amsthm}
\usepackage{graphicx,booktabs}
\newtheorem{proposition}{Proposition}
\DeclareMathOperator{\norm}{norm}
\newcommand{\R}{\mathbb{R}}

\newcommand{\StandardFull}{\ensuremath{64.60}}
\newcommand{\StandardLoopFull}{\ensuremath{64.14}}
\newcommand{\StandardFullLoss}{\ensuremath{0.46}}
\newcommand{\RegisterFullLoss}{\ensuremath{0.77}}
\newcommand{\FullTuneFullLoss}{\ensuremath{0.76}}
\newcommand{\GenerativeFullDelta}{\ensuremath{+0.36}}
\newcommand{\NativeFullDelta}{\ensuremath{+4.23}}
\newcommand{\NativeFull}{\ensuremath{43.19}}
\newcommand{\StandardSecondPassGain}{\ensuremath{1.27}}

\newcommand{\GenerativeSeedZeroDelta}{\ensuremath{+0.66}}
\newcommand{\GenerativeSeedOneDelta}{\ensuremath{+0.62}}
\newcommand{\GenerativeWithoutChart}{\ensuremath{0.14}}
\newcommand{\GenerativeQueryDelta}{\ensuremath{+0.15}}
\newcommand{\StandardFreezeDelta}{\ensuremath{+0.15}}
\newcommand{\BidirectionalFreezeDelta}{\ensuremath{+0.12}}
\newcommand{\AlternatingFreezeDelta}{\ensuremath{+0.21}}
\newcommand{\GenerativeFreezeDelta}{\ensuremath{+0.08}}
\newcommand{\AlternatingSensitivity}{\ensuremath{0.044}}
\newcommand{\AlternatingSensitivityMatched}{\ensuremath{0.043}}
\newcommand{\OrderImageFirst}{\ensuremath{67.19}}
\newcommand{\OrderQueryFirst}{\ensuremath{65.82}}
\newcommand{\OrderSandwich}{\ensuremath{67.43}}
\newcommand{\OrderBidirectional}{\ensuremath{67.94}}
\newcommand{\SandwichQueryDelta}{\ensuremath{+0.24}}
\newcommand{\BidirectionalQueryDelta}{\ensuremath{+0.75}}

\newcommand{\NativeEqualStepNative}{\ensuremath{42.25}}
\newcommand{\NativeEqualStepSubset}{\ensuremath{35.10}}
\newcommand{\NativeMatchedSubset}{\ensuremath{38.70}}
\newcommand{\NativeMatchedDelta}{\ensuremath{+3.35}}

\newcommand{\NativeFreezeLoss}{\ensuremath{10.42}}
\newcommand{\NativeTextFreezeLoss}{\ensuremath{10.25}}
\newcommand{\SwapKnowledgeLoss}{\ensuremath{11.10}}
\newcommand{\SwapInfographicLoss}{\ensuremath{11.70}}
\newcommand{\SwapGroundingLoss}{\ensuremath{9.80}}
\newcommand{\SwapCirrDelta}{\ensuremath{+0.10}}
\newcommand{\SwapFashionDelta}{\ensuremath{-1.40}}

\definecolor{cvprblue}{rgb}{0.21,0.49,0.74}
\usepackage[pagebackref,breaklinks,colorlinks,allcolors=cvprblue]{hyperref}
\title{When Does Reusing Layers Help Multimodal Embeddings?}
\author{Zhizhang Hu\thanks{This work is independent of, and unrelated to, Amazon.}\\
Amazon\\
{\tt\small zhizhanh@amazon.com}
\and
Yue Zhang\\
University of California, Merced\\
{\tt\small yzhang58@ucmerced.edu}}
\begin{document}
\maketitle
\begin{abstract}
We ask whether a compact universal multimodal embedding models can produce better search embeddings by running the same layers again.
Looped transformers add computation with nearly unchanged model size, making them an appealing alternative to a larger encoder.
Their success in language-model reasoning and recent use for retrieval motivate the idea, but higher retrieval scores alone do not reveal what an extra pass contributes.
We study eleven loop designs and training variants on three checkpoints of one vision--language architecture.
Our protocol distinguishes whether a model uses its extra pass, whether its image states respond to the query, and whether those updates improve retrieval.
We prove that an image placed before the query remains query-independent through repeated layers when information only travels forward.
We then open a query-to-image path in three ways, including order-alternating recurrence, which places the query first on the next pass.
Each path produces query-sensitive image states.
For bidirectional attention and order alternation, freezing later visual updates does not lower accuracy.
On all 78 datasets of the Massive Multimodal Embedding Benchmark version 2 (MMEB-V2), a standard added loop loses \StandardFullLoss{} points against its trained control; an intermediate generative stage changes the effect to \GenerativeFullDelta{}.
A model pretrained with recurrence gains \NativeFullDelta{} points over its own loop-removed control and benefits from query-conditioned image updates when answering visual questions or locating objects from text.
\end{abstract}

\section{Introduction}
\label{sec:intro}

A universal multimodal embedding model compresses text, images, videos, or document pages into vectors~\cite{zhou2026wemm}.
The following retrieval and grounding then becomes a comparison between a query vector and candidate vectors.
This lets one model support tasks as different as finding a photograph from a description, locating evidence in a document, and answering a visual question by retrieving the right answer~\cite{jiang2024vlm2vec,meng2025vlm2vecv2}.
Empirically, larger models often provide stronger representations, but also require higher-end GPUs with more memory, which hinders the scaling up for industry-level applications.
In this work, we ask, \emph{can a small model improve its embeddings quality by reusing its existing layers instead of acquiring more parameters?}

A looped transformer applies the same layer block several times to an evolving hidden state.
It increases executed depth while sharing the block's weights.
Depth recurrence has a long history~\cite{dehghani2019universal,lan2020albert}, and recent large language models (LLMs) use it to improve reasoning without generating a longer chain of text~\cite{geiping2025huginn,zhu2025ouro}.
For multimodal retrieval, ReLoop-UME proposes to reuse learned retrieval registers and reports a gain from repeated depth~\cite{wang2026reloop}.
This makes looping promising, but leaves a practical question: what does the extra pass contribute to the universal multimodal embedding model?

The query provide a concrete reason to revisit the hidden state.
A same bus photograph can support two different queries of ``When is this scene taking place?'' and ``How many windows are visible?''
A second pass could represent the image differently after reading the question.
However, most causal embedding prompts put image tokens before the instance text.
Information travels forward in that order: the text can read the image, while the image cannot read the text.
Repeating the same causal order preserves this restriction.
Even though the final embedding can still depend on both inputs, what is unavailable is a query-conditioned update of the image states inside the language stack.
We denote such an update visual re-encoding.

We investigate this opportunity and its limits with Qwen3.5-2B~\cite{qwen2026qwen35} and two universal multimodal embedding built on the same architecture, UEmbed~\cite{song2026uembed} and WeChat Multimodal Embedding~\cite{zhou2026wemm}.
Our study covers eleven loop variants, spanning continued embedding training, contrastive conversion, query information paths, full decoder fine-tuning, and an intermediate generative training stage.
For each variant we compare with a separately trained control.
We additionally convert the natively recurrent LoopVL-1B model~\cite{qian2026loopvl} into an universal multimodal embedding model and compare it with its own loop-removed counterpart.

\begin{figure*}[t]
\centering
\includegraphics[width=\textwidth]{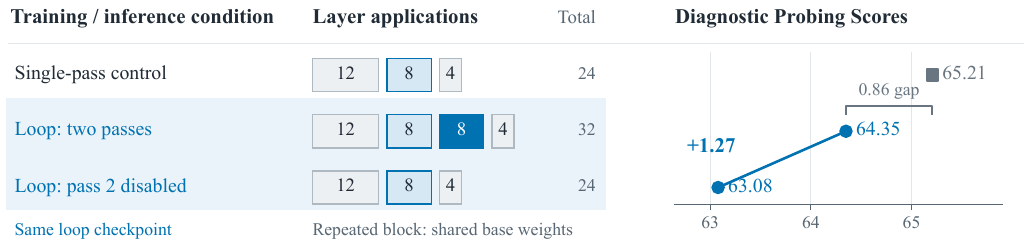}
\caption{\textbf{Using an extra pass is different from improving the model.}
On the 16-dataset diagnostic, the same trained Qwen3.5-2B loop improves by \StandardSecondPassGain{} points when its second pass is enabled.
A separately trained single-pass control still scores higher.
The middle and right points use exactly the same loop checkpoint.
Boxes count layer applications, with the repeated eight-layer block sharing its base weights.
The plotted score axis is restricted to show the differences.}
\label{fig:comparison}
\end{figure*}

Our analysis probes three aspects: reliance on extra computation, query sensitivity, and benefit from visual updates. 
We control the first aspect by disabling a pass, the second aspect by changing the query while holding the image fixed, and the third aspect via intervening on later visual states. 
We also investigate varying the information path by introducing order-alternating recurrence: an image-first pass is followed by a query-first pass, after which the states return to their original order. 
This permits two-way image–text exchange through causal layers, including recurrent sequence layers, without duplicating tokens or adding parameters for the permutation.
Together with bidirectional attention and pooled-state feedback, it lets us test whether opening a query-to-image path is enough.

On the full MMEB-v2 benchmark, standard recurrence falls below its trained single-pass control. 
Adding generative adaptation yields a small positive loop effect, concentrated largely in chart-document retrieval.
The native LoopVL recurrence gives a larger gain over its loop-removed control, including with approximately matched conversion time. 
For the information flow, the query-conditioned visual computation is task-dependent with no uniformally one-fit-all recipe.

Our contributions includes: (1). a controlled comparison of added and native recurrence for universal multimodal embedding models, (2). a causal analysis and token-order intervention that separate depth from information flow, and (3) state interventions that distinguish query sensitivity from useful visual refinement.
We envision this work lays the foundation for future advancements on looped transformer-based universal multimodal embedding.

\section{Related work}
\label{sec:related}

\paragraph{Reusing depth.}
Universal Transformers recurrently apply a shared transition~\cite{dehghani2019universal}; cross-layer parameter sharing also reduces encoder size~\cite{lan2020albert}.
Huginn and Ouro, recurrent language models trained with shared depth blocks, show that latent iterations can improve reasoning~\cite{geiping2025huginn,zhu2025ouro}.
Controlled design studies further distinguish where recurrence is placed, how input features are injected, and how recurrent states are supervised~\cite{huang2026loopedI,huang2026loopedII}.
Recurrence can also be added after pretraining: McLeish et al.\ use a depth curriculum when adapting language models~\cite{mcleish2025retrofitted}.
These results motivate our experiments, but success in next-token prediction does not establish success under supervision on a single retrieval vector.
Our generative retrofit tests one fixed-depth adaptation, not the full range of such curricula.

\paragraph{Multimodal embeddings.}
Vision--Language Model to Vector (VLM2Vec) converts instruction-following multimodal models through contrastive training~\cite{jiang2024vlm2vec}; its second version extends evaluation to video and visual documents~\cite{meng2025vlm2vecv2}.
Qwen3-VL-Embedding, a retrieval encoder based on the Qwen3 vision--language family, uses staged contrastive training and distillation~\cite{li2026qwen3vlembedding}.
UEmbed produces dense and sparse representations in one causal pass~\cite{song2026uembed}, while WeMM-Embedding combines alignment, relevance refinement, and cross-scale knowledge transfer~\cite{zhou2026wemm}.
We study their \emph{dense} embeddings.
ReLoop-UME is the closest depth-recurrent method: it repeats a middle-to-late block and uses five learned registers as persistent retrieval states~\cite{wang2026reloop}.
On Qwen3.5-2B it reports 60.2 to 62.6 on the full benchmark.
Our register-based test follows its architecture under our own conversion recipe.

\paragraph{Making representations depend on the query.}
Bidirectional attention can turn a causal text model into an encoder, as in Large Language Model to Vector (LLM2Vec)~\cite{behnamghader2024llm2vec}.
Echo embeddings repeat text so its later states can access the entire earlier copy~\cite{springer2024echo}; Just Read Twice similarly addresses order-dependent recall in recurrent sequence models~\cite{arora2024jrt}.
For vision, instruction-aware feature extraction and a Question Aware Vision Transformer already condition visual features on text~\cite{dai2023instructblip,ganz2024qavit}.
LoopVL combines recurrent multimodal states with a bidirectional input prefix and repeated visual-feature injection~\cite{qian2026loopvl}.
Our order alternation changes the \emph{hidden-state order between depth passes}, rather than duplicating input tokens or modifying the vision encoder.
Its purpose here is to expose the query information path and test whether the embedder makes useful use of it.

\section{Depth, information flow, and useful updates}
\label{sec:method}

\subsection{A looped multimodal embedder}
\label{sec:embedder}

An embedder maps an input $x$ to a unit vector $f(x)\in\R^d$.
Queries and candidates share the encoder and are ranked by cosine similarity.
The contrastive objective increases the similarity of a matched query--candidate pair relative to other candidates in the batch~\cite{jiang2024vlm2vec}.
We use UEmbed's prompt and dense readout~\cite{song2026uembed}.
The task instruction is separate from the instance text, and the representation is pooled at the last token of the generation prompt.
We write the four input segments as
\begin{equation}
x=(c,v,q,t),
\label{eq:segments}
\end{equation}
where $c$ is context and the task instruction, $v$ contains the visual tokens, $q$ is the instance question, modification, or description, and $t$ is the tail containing the pooling position.
The frozen vision encoder supplies the initial visual features; $v$ refers to their states inside the language decoder.

The main base model used is Qwen3.5-2B. It has 24 layers: six groups of three Gated DeltaNet layers, which update a causal sequence state, followed by one full-attention layer~\cite{qwen2026qwen35,yang2024gateddeltanet}.
Our standard loop uses a prelude $P$ of 12 layers, a repeated block $R$ of 8 layers, and a coda $C$ of 4 layers.
The same base weights implement each use of $R$.
With $r$ passes, the state is
\begin{align}
s_1&=R(P(x)), \nonumber\\
s_k&=s_{k-1}+g_k\odot\big(R_k(s_{k-1})-s_{k-1}\big),
\quad k=2,\ldots,r,
\label{eq:recurrence}
\end{align}
and $f(x)=\norm(C(s_r)_{\mathrm{pool}})$.
Here $\norm$ denotes unit-length normalization.
Each $R_k$ has a pass-specific low-rank adaptation (LoRA)~\cite{hu2021lora}, and $g_k\in\R^d$ is a learned elementwise gate.
A zero gate suppresses the additional pass; a unit gate gives plain repetition.
The standard $r=2$ loop executes 32 rather than 24 layers and adds fewer than 1\% parameters.

\subsection{What repeated causality cannot change}
\label{sec:causality}

\begin{proposition}[Image states remain query-independent]
\label{prop:causal}
Suppose each layer is causal in the same position order: an output at position $i$ depends only on inputs at positions $j\leq i$.
Suppose gates, adapters, and input injection preserve that dependency.
For the image-first sequence $(c,v,q,t)$, every visual-token state is independent of $q$ after every layer of every depth pass.
\end{proposition}

\noindent\emph{Proof.}
Composing causal maps preserves causality.
Equation~\eqref{eq:recurrence} mixes such maps positionwise, also preserving causality.
All positions of $q$ follow all positions of $v$.

The proposition covers causal attention, Gated DeltaNet scans, causal convolutions, and positionwise feed-forward layers.
It assumes fixed context and image preprocessing; a feedback term derived from later query positions would violate its premise.
The result restricts \emph{where} query information can travel, not how much retrieval can improve.
Text and pooling states can reread visual evidence on every pass, and a loop may improve retrieval through those states alone.
Learned registers appended after the input also remain downstream of the image: repeating them does not by itself create a reverse information path.

\begin{figure*}[t]
\centering
\includegraphics[width=\textwidth]{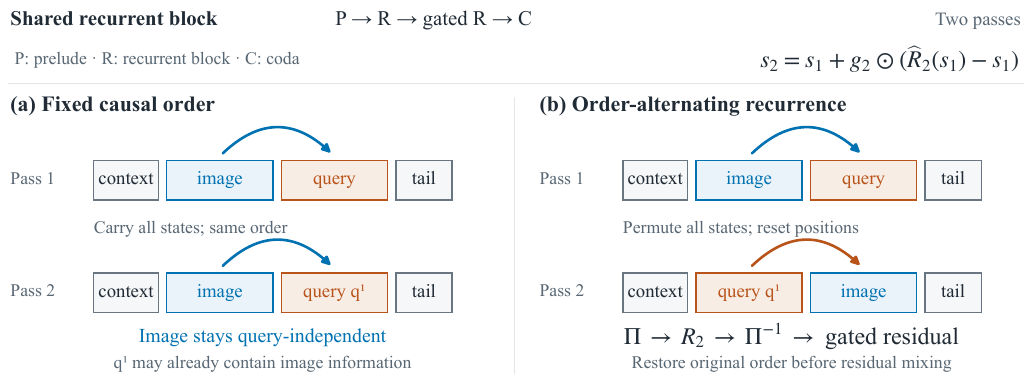}
\caption{\textbf{More depth does not automatically change who can read whom.}
Arrows show possible information flow, with the arrowhead at the reader.
Both passes in the fixed image-first order let the query read the image.
Order-alternating recurrence places the already image-informed query before the image on pass 2, opening a reverse path through the same causal block.
States and positional indices are permuted together, and the output is restored before the gated residual.
The prelude and coda run once; only the middle block is reused.
Opening a path permits dependence but does not guarantee that training will exploit it.}
\label{fig:flow}
\end{figure*}

\subsection{Three ways to open a query-to-image path}
\label{sec:channels}

\paragraph{Bidirectional attention.}
We remove the causal mask from the six full-attention layers.
This lets visual states read later query states, following the text-encoder conversion used by LLM2Vec~\cite{behnamghader2024llm2vec}.
The 18 Gated DeltaNet layers still scan causally; changing an attention mask does not reverse those scans.
A separate single-pass control uses the same bidirectional attention.

\paragraph{Pooled-state feedback.}
The pooling state has already seen both the image and query.
We apply root-mean-square normalization (RMSNorm), project through rank-64 matrices, and add the result to every position before the next pass:
\begin{equation}
\widetilde{s}_{k-1}[i]
=s_{k-1}[i]+B_k A_k\,\mathrm{RMSNorm}(s_{k-1}[\mathrm{pool}]).
\label{eq:feedback}
\end{equation}
$B_k$ starts at zero.
This sends a summary of the query back to the visual positions without changing sequence order.

\paragraph{Order-alternating recurrence.}
Instead of changing a layer, we change the order of its hidden-state inputs.
Odd passes use $(c,v,q,t)$.
Even passes gather the states into $(c,q,v,t)$, run the causal block, and restore the original order before applying the residual in Eq.~\eqref{eq:recurrence}.
If $\Pi_k$ is the permutation, the repeated branch is
\begin{equation}
\widehat R_k(s)=\Pi_k^{-1}R_k(\Pi_k s;\,p_k),
\label{eq:alternation}
\end{equation}
with positional indices $p_k$ recomputed for the permuted sequence.
Multimodal rotary position embeddings are reindexed using the model's native position assignment~\cite{wang2024qwen2vl}.
No tokens are duplicated, and no parameters are added by the permutation.
Inputs lacking either the visual or instance-text segment retain their original order.

After pass 1, each query state can contain information about the image.
In pass 2, visual states can read these query states, so they can depend on both the image and the full query.
Query states can retain their earlier image dependence through the residual stream, while their within-query dependence remains causal.
Two passes therefore permit two-way exchange at the segment level; they do not make every token attend to every other token.

\paragraph{A one-pass comparison.}
We also train a query-sandwich control with duplicated instance text, $(c,q,v,q,t)$.
The first query copy informs the image; the second reads the image.
This provides two-way exchange within one causal pass, at the cost of $|q|$ extra tokens, paralleling input repetition in text embeddings~\cite{springer2024echo}.
It tests the value of the information path independently of recurrent depth.

\subsection{Separating sensitivity from benefit}
\label{sec:diagnostics}

We use three complementary comparisons.
First, a trained single-pass control tests whether the loop improves a model trained for the same retrieval task.
Second, removing the extra pass at inference tests whether a trained loop relies on that pass.
These comparisons differ because the loop was optimized to be read out after its final pass.

Third, we intervene on later-pass visual states.
Visual freeze holds them at their incoming values throughout the repeated block, while text and readout states continue to update.
The change in retrieval score tests dependence on the visual updates, although such a freeze can also disrupt the recurrent computation.
For the native loop we add a same-image query swap: visual states are computed from the same image with another query, then substituted into the original run while its actual text states remain.
This targets the query-specific part of visual computation more directly than freezing.
A text-side freeze and a blocked visual-carry intervention supply additional controls.

We measure query sensitivity separately:
\begin{equation}
S_k=\mathbb E_{(v,q,q')}
\frac{\|h_v^{(k)}(v,q)-h_v^{(k)}(v,q')\|_2}
{\|h_v^{(k)}(v,q)\|_2}.
\label{eq:sensitivity}
\end{equation}
The image and task instruction stay fixed; $q'$ comes from another example.
A length-matched check controls positional shifts under alternation.
$S_k>0$ establishes dependence on the query, but usefulness requires a retrieval effect under an intervention.

\section{Experimental setup}
\label{sec:setup}

\paragraph{Contrastive conversion.}
We select Qwen3.5-2B as our main experimental backbone model for its light-weight yet strong performance~\cite{qwen2026qwen35}. Given it is pretrained to generate text, we first adopt a contrastive conversion which trains its pooled representations for embedding retrieval.
Our main comparisons start from Qwen3.5-2B and train single-pass and added-loop models for 3,000 steps with task-consistent batches.
The data are 20 image-centered training datasets from the original Massive Multimodal Embedding Benchmark (MMEB), totaling 1.07 million query--candidate pairs~\cite{jiang2024vlm2vec}.
We use LoRA on all decoder linear layers, keeping the vision encoder and merger frozen.
Two-pass variants add rank-16 pass-specific LoRA and unit-initialized gates.
A separate standard-loop pair trains all decoder layer weights.

Within each added-loop comparison, training length, data mixture, preprocessing, and evaluation are fixed.
We adopt matched training seeds to fixe adapter initialization, source dataset, and batch size at each step.
We in addition experiment with two VLM2Vec-style embedding mdoels: the UEmbed~\cite{zhou2026wemm} and WeMM-Embedding~\cite{zhou2026wemm}. Given they are already produce retrieval embeddings, so we continue embedding training from their released weights.
We cover the set of of native-looped model, i.e., LoopVL-1B and its loop-removed control undergo conversion in Section~\ref{sec:native-results}.

\paragraph{Generative retrofit.}
We also test whether image-grounded generative supervision changes the value of an added loop.
Starting from Qwen3.5-2B, we train an order-alternating two-pass model and a single-pass control with answer-token cross-entropy for 6,000 steps on 733,000 examples.
The sources are 13 image-to-text datasets from the embedding-training mixture and first-turn LLaVA instruction data~\cite{liu2023llava}.
Both models then undergo the same 3,000-step contrastive conversion, so generative supervision precedes the pooled embedding objective.

\paragraph{Full benchmark and diagnostic subsets.}
We evaluate the image-only adaptation on the broader MMEB-v2~\cite{meng2025vlm2vecv2}, whose 78 datasets comprise 36 image, 18 video, and 24 visual-document datasets.
We report official top-1 hit rate for image and video tasks, and normalized discounted cumulative gain at 5 (nDCG@5) for documents, on a 0--100 scale.
The overall score averages all dataset scores, weighting families in proportion to their dataset counts.

For repeated runs and state interventions, we use a 16-dataset subset of MMEB-v2: 8 image, 4 video, and 4 document datasets.
Six of these contain both an image and instance text in one encoded input, allowing us to test query-conditioned visual updates.
The mixed inputs are queries in A-OKVQA~\cite{schwenk2022aokvqa}, InfographicsVQA, RefCOCO~\cite{lin2014coco}, CIRR~\cite{liu2021cirr}, and FashionIQ~\cite{wu2021fashioniq}, and image--description candidates in EDIS~\cite{liu2023edis}.
The mean of these six dataset scores is the \emph{query-conditioned score}; the other ten form a complementary mean.

\paragraph{Full-gallery composed retrieval.}
We additionally test composed image retrieval with full validation galleries to check whether the findings hold beyond MMEB-v2's sampled candidate pools.
CIRR and FashionIQ reuse benchmark datasets under this protocol; CIRCO adds a dataset with multiple relevant targets.
These scores are separate from the MMEB-v2 aggregate.
CIRR uses 4,181 queries and 2,297 images, excludes the reference image, and reports recall at 1 (R@1).
FashionIQ reports recall at 10 (R@10), averaged across three clothing categories.
CIRCO~\cite{baldrati2023searle} uses 220 queries and 123,403 images, reporting mean average precision at 5 (mAP@5).

\paragraph{Repeated runs.}
The main standard, bidirectional, alternating, generative, and native comparisons have two training random seeds.
Register recurrence, feedback, full fine-tuning, and released-embedder tests have one.
Full-benchmark results use the seed-0 training checkpoints.
We report both seed-matched differences instead of treating two seeds as a precise estimate of variance.

\section{What the extra passes contribute}
\label{sec:results}

\subsection{Added loops give limited retrieval gains}
\label{sec:added-results}

\begin{table*}[t]
\centering
\small
\begin{tabular}{@{}lrrrrr@{}}
\toprule
Model and training & Image & Video & Document & Overall & $\Delta$ \\
& (36) & (18) & (24) & (78) & paired control \\
\midrule
\multicolumn{6}{@{}l}{\textit{Qwen3.5-2B: loops added after pretraining}} \\
Standard conversion & 69.58 & 51.03 & 67.30 & 64.60 & \textemdash \\
\quad + standard causal loop & 69.69 & 50.48 & 66.06 & 64.14 & $-0.46$ \\
\quad + pooled-state feedback & 69.78 & 51.00 & 65.86 & 64.24 & $-0.36$ \\
\quad + order-alternating recurrence & 69.51 & 50.90 & 65.70 & 64.04 & $-0.56$ \\
\addlinespace[2pt]
Register-equipped conversion & 69.93 & 50.90 & 66.04 & 64.34 & \textemdash \\
\quad + register recurrence (four passes) & 69.65 & 49.41 & 65.07 & 63.57 & $-0.77$ \\
\addlinespace[2pt]
Generative retrofit + conversion & 69.10 & 48.99 & 64.39 & 63.01 & \textemdash \\
\quad + order-alternating recurrence & 69.04 & 50.02 & 64.87 & 63.37 & $+0.36$ \\
\addlinespace[2pt]
Full decoder fine-tuning & 69.04 & 47.00 & 66.62 & 63.21 & \textemdash \\
\quad + standard causal loop & 68.54 & 46.15 & 65.52 & 62.45 & $-0.76$ \\
\midrule
\multicolumn{6}{@{}l}{\textit{LoopVL-1B: recurrent pretrained weights}} \\
Native model, loop removed, longer training & 50.03 & 25.96 & 32.12 & 38.96 & \textemdash \\
\quad Native recurrence & 51.72 & 29.99 & 40.28 & 43.19 & $+4.23$ \\
\bottomrule
\end{tabular}
\caption{\textbf{Complete 78-dataset evaluation.} Each difference uses the control in its own training group. The overall score averages all datasets, rather than averaging the three family means. The native loop-removed control trains longer to approximately match observed conversion time. These two model groups are not an architecture-matched comparison. Bidirectional and query-sandwich variants are not assigned an unmeasured full-benchmark score.}
\label{tab:full-benchmark}
\end{table*}

Table~\ref{tab:full-benchmark} illustrates the full 78-dataset results.
The standard single-pass conversion reaches \StandardFull{}, versus \StandardLoopFull{} for its two-pass loop variant.
The register-based loop loses \RegisterFullLoss{} points against its own register-equipped control.
Pooled feedback and order alternation also fall below the standard control.
Training all decoder layers still leaves a \FullTuneFullLoss{}-point loop deficit.

Repeated diagnostic runs show seed variation (Table~\ref{tab:paired}): bidirectional and alternating loops both change sign.
UEmbed and WeMM-Embedding loops fall below their continued controls in single runs.

\begin{table}[t]
\centering
\small
\begin{tabular}{@{}lrrr@{}}
\toprule
Loop setting & Seed 0 & Seed 1 & Mean \\
\midrule
Standard causal loop & $-0.86$ & $-0.09$ & $-0.47$ \\
Bidirectional attention & $-0.39$ & $+0.44$ & $+0.02$ \\
Order alternation & $-0.68$ & $+0.17$ & $-0.26$ \\
Generative retrofit & $+0.66$ & $+0.62$ & $+0.64$ \\
Native recurrence & $+3.42$ & $+3.28$ & $+3.35$ \\
\bottomrule
\end{tabular}
\caption{\textbf{Repeated diagnostic comparisons.} Loop minus its trained control on the 16-dataset mean, in points. The bidirectional control has the same attention change; the generative control has the same extra training stage. The native control has approximately matched observed training time.}
\label{tab:paired}
\end{table}

\paragraph{The positive retrofit exception.}
Generative training changes the loop comparison: its two diagnostic gains are \GenerativeSeedZeroDelta{} and \GenerativeSeedOneDelta{} points, and its full-benchmark gain is \GenerativeFullDelta{}.
Most of the diagnostic gain comes from chart questions: removing that document task leaves \GenerativeWithoutChart{} points, while the query-conditioned mean improves by \GenerativeQueryDelta{}.
Both generatively trained models remain below the ordinary contrastive conversion.
The retrofit changes the marginal loop effect without improving the strongest overall model.
Visual freezing has no accuracy cost, so the gain does not establish useful query-conditioned visual updates.

\paragraph{How this relates to ReLoop-UME.}
The published Qwen3.5-2B result is 60.2 for one pass and 62.6 for ReLoop-UME~\cite{wang2026reloop}.
Our single-pass score of \StandardFull{} comes from a stronger conversion recipe.
ReLoop-UME trains the language model, vision encoder, and merger on 24 image, video, and document sources for 5,000 steps; our standard conversion uses image-only data, frozen vision, and LoRA for 3,000 steps.
Our reimplementation tests architectural transfer to this recipe, yielding a loss without repeating the published training procedure.

\subsection{A loop can be active without useful visual refinement}
\label{sec:mechanism-results}

\begin{figure*}[t]
\centering
\includegraphics[width=.9\textwidth]{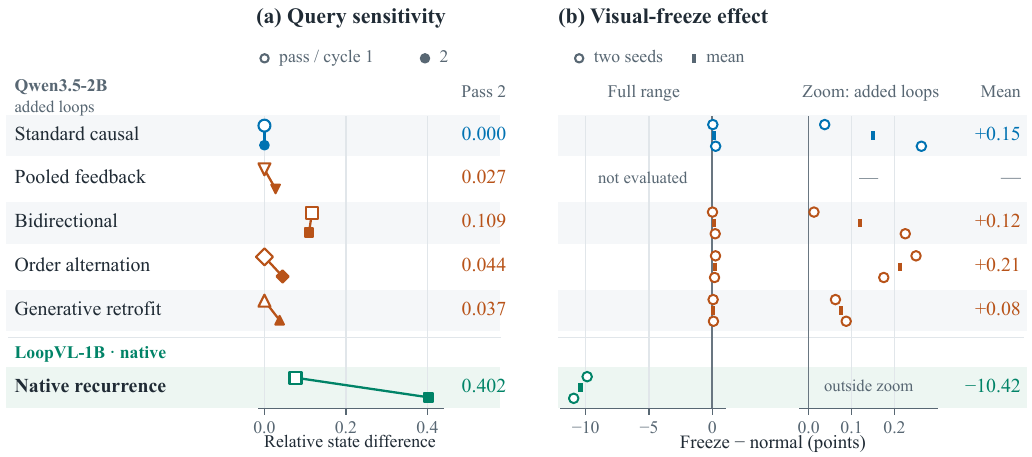}
\caption{\textbf{Query dependence and useful visual updates are separate properties.}
Left: dataset-averaged relative visual-state differences after pass or cycle 1 and 2 when only the query changes (one checkpoint).
This measures hidden states, not retrieval accuracy.
Right: changes in the eight-image-task score under visual freeze; circles show seed-matched changes and bars their means.
Opening a query path produces sensitivity without an accuracy cost under freeze for the measured added loops.
The native loop is strongly affected; Figure~\ref{fig:swap} tests whether its dependence is query-specific.}
\label{fig:mechanism}
\end{figure*}

Removing pass 2 at inference lowers the standard loop's diagnostic score by \StandardSecondPassGain{} points.
Query-aware variants also lose accuracy after one pass, showing reliance on extra computation.
Yet visual freeze changes the eight-image-task mean by \StandardFreezeDelta{} for the standard loop, \BidirectionalFreezeDelta{} for bidirectionality, \AlternatingFreezeDelta{} for alternation, and \GenerativeFreezeDelta{} after the retrofit.
All four changes are small and positive over two seeds, so the extra pass can operate through text and readout computation without requiring those visual updates.

\begin{figure*}[t]
\centering
\includegraphics[width=.9\textwidth]{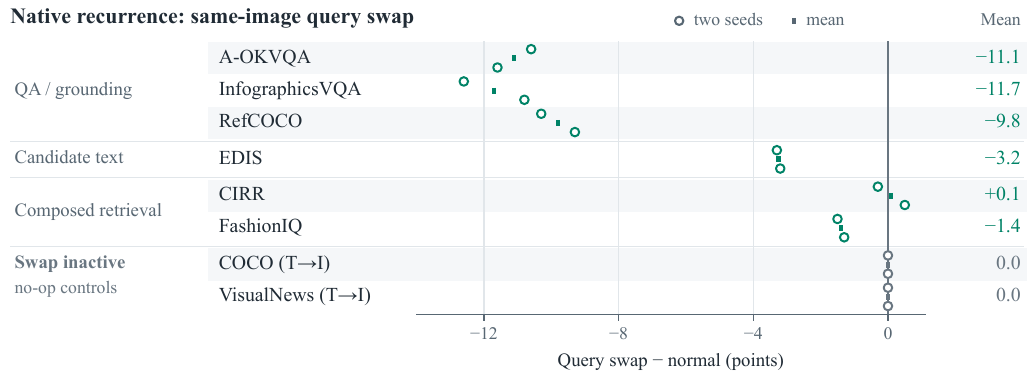}
\caption{\textbf{Which tasks need the image to be updated under the right query?}
In the native LoopVL embedder, second-cycle visual states are replaced by states computed from the same image under another instance text; actual text states are retained.
Dots show two seeds and short bars their means.
Scores use MMEB's diagnostic candidate pools.
Question answering and grounding are more affected than composed retrieval.
Generic text-to-image inputs provide a zero-change control.}
\label{fig:swap}
\end{figure*}

\begin{figure}[t]
\centering
\includegraphics[width=.85\columnwidth]{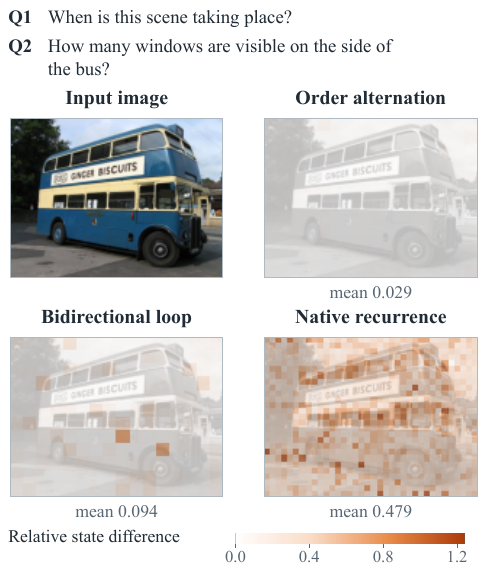}
\caption{\textbf{The same image, two questions.}
Maps show per-patch relative changes in visual hidden states after pass or cycle 2, using a prespecified Visual7W example~\cite{zhu2016visual7w}.
Models use native grids on a shared, unclipped scale.
These are state-difference maps, not attention or saliency maps.
}
\label{fig:qual}
\end{figure}

\begin{figure*}[t]
\centering
\includegraphics[width=\textwidth]{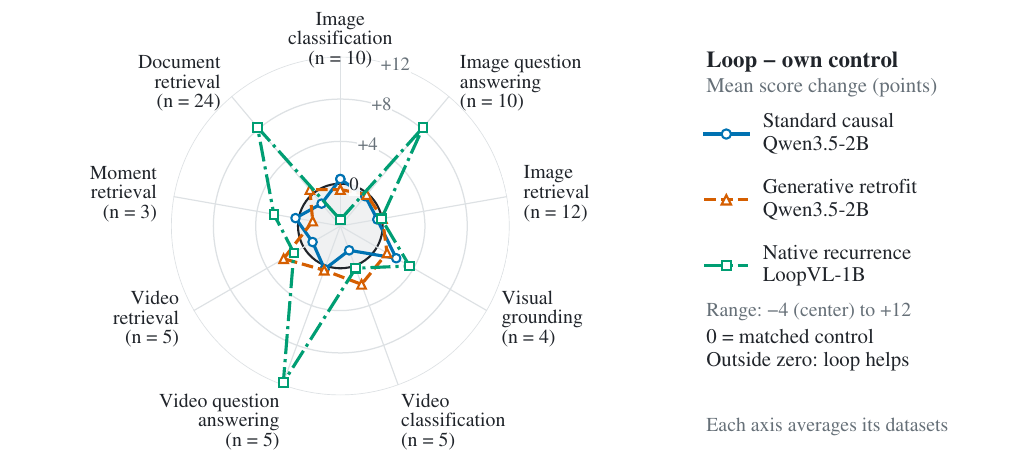}
\caption{\textbf{Which task groups benefit from recurrence?}
Axes partition all 78 MMEB-v2 datasets; points average dataset-score differences from each model's own trained control.
Scores use top-1 hit rate for images and videos and nDCG@5 for documents, on a 0--100 scale.
The retrofit uses order-alternating recurrence.
The native control uses the same pretrained weights without repetition and trains longer.
The zero ring separates gains from losses.}
\label{fig:task-profile}
\end{figure*}

Figure~\ref{fig:mechanism} tests the information path directly.
The image-first causal loop has exactly zero query sensitivity in both passes, as Proposition~\ref{prop:causal} predicts.
Feedback and alternation acquire sensitivity on pass 2; bidirectional attention has it on pass 1, without an increase on pass 2.
Length-matching changes alternating sensitivity from \AlternatingSensitivity{} to \AlternatingSensitivityMatched{}, so visual-token position shifts explain little of the effect.
Bidirectional and alternating loops have no accuracy penalty under visual freeze; feedback freeze was not evaluated.

The bus-question maps in Figure~\ref{fig:qual} show hidden-state changes; usefulness is tested by the retrieval interventions.

\paragraph{Token order without a loop.}
The single-pass image-first, query-first, query-sandwich, and bidirectional controls have query-conditioned scores of \OrderImageFirst{}, \OrderQueryFirst{}, \OrderSandwich{}, and \OrderBidirectional{}, respectively.
Simply moving the query first loses accuracy: it lets the image read the query while preventing the query from reading the image.
Duplicating the query yields a small \SandwichQueryDelta{}-point gain; bidirectional attention yields \BidirectionalQueryDelta{}.
Duplicated long video questions exceed the input limit, so the sandwich's full-benchmark evaluation is incomplete and its overall score is omitted.

\subsection{Native recurrence provides a contrasting case}
\label{sec:native-results}

LoopVL-1B has two distinct 16-layer stacks, denoted $L$ and $H$.
Its released schedule, $3L1H$, has two outer cycles and executes 128 layers from 32 unique layers~\cite{qian2026loopvl}.
Its entire embedding input is a bidirectional prefix, with fixed visual features reinjected at each cycle.
We convert it with rank-32 LoRA for 1,500 steps at global batch size 128, keeping vision and the projector frozen.

Our loop-removed control runs $L\,H$ using the \emph{same pretrained weights}.
At equal conversion steps, native recurrence scores \NativeEqualStepNative{} versus \NativeEqualStepSubset{} on the diagnostic subset.
The native schedule takes 25.3\,s per step versus 6.7\,s for the loop-removed schedule.
Allowing the latter 5,700 steps approximately matches the observed training time and raises its score to \NativeMatchedSubset{}.
Native recurrence retains a \NativeMatchedDelta{} diagnostic gain (paired runs in Table~\ref{tab:paired}) and gains \NativeFullDelta{} on all 78 datasets.
Its absolute full-benchmark score, \NativeFull{}, is well below the Qwen3.5 control.

\paragraph{Visual dependence is not yet query-specific.}
Freezing second-cycle visual states lowers the native model's eight-image-task mean by \NativeFreezeLoss{} points.
Freezing text states, except the readout, lowers it by \NativeTextFreezeLoss{}.
Similar losses show disruption on both sides of the recurrence, so visual freeze alone cannot identify query-specific usefulness.

The same-image query swap lowers A-OKVQA, InfographicsVQA, and RefCOCO by \SwapKnowledgeLoss{}, \SwapInfographicLoss{}, and \SwapGroundingLoss{} points (Fig.~\ref{fig:swap}), showing dependence on visual states computed under the actual question or referring expression.
Composed retrieval is less affected: \SwapCirrDelta{} on diagnostic CIRR and \SwapFashionDelta{} on FashionIQ despite large freeze losses.
These tests provide limited evidence that its second-cycle benefit comes from query-specific image updates.

\subsection{Task-wise gains depend on the training regime}
\label{sec:task-profile}

The full-benchmark profiles in Figure~\ref{fig:task-profile} show a 2.10-point grounding gain for the standard loop, alongside losses of 1.55 on video classification and 1.24 on document retrieval.
After the retrofit, the loop gains 2.19 on video retrieval and 1.87 on video classification, but loses 1.35 on moment retrieval.
Native recurrence gains most on video question answering (+11.75), image question answering (+8.16), and document retrieval (+8.17), while image classification declines by 3.41.
The grounding gain coexists with query-independent visual states, illustrating why task-wise gains need not come from query-conditioned visual updates.

\section{Conclusion}
\label{sec:conclusion}
Through extensive controlled experiments, we find extra computation and query-sensitive image states do not by themselves establish better embeddings.
However, the added loops provide limited, task-dependent gains under the tested recipes, including a small positive generative effect.
To fully unleash the efficacy of looped transformer, the multimodal embedding model needs to be converted from a model pre-trained with native recurrence fashion. This helps against its own loop-removed weights, most in question answering and document retrieval.
Query swaps show that the correct query-conditioned visual states matter for question answering and grounding.
We envision the evidence from the extensive research will lay a solid foundation for the future investigation of the looped transformer-based universal embedding and understanding.

\clearpage
\section*{LLM Usage Statement}
Large language models (LLMs) are used throughout the project.
Anthropic's Claude Opus 5.5 model assists with literature review and experiment implementation.
OpenAI's GPT-6.1-Sol model assists with manuscript's language and presentation quality check (e.g., spelling, grammar, etc.)

{\small
\bibliographystyle{ieeenat_fullname}
\bibliography{main}

\begin{thebibliography}{31}
\providecommand{\natexlab}[1]{#1}
\providecommand{\url}[1]{\texttt{#1}}
\expandafter\ifx\csname urlstyle\endcsname\relax
  \providecommand{\doi}[1]{doi: #1}\else
  \providecommand{\doi}{doi: \begingroup \urlstyle{rm}\Url}\fi

\bibitem[Arora et~al.(2024)Arora, Timalsina, Singhal, Spector, Eyuboglu, Zhao,
  Rao, Rudra, and R{\'e}]{arora2024jrt}
Simran Arora, Aman Timalsina, Aaryan Singhal, Benjamin Spector, Sabri Eyuboglu,
  Xinyi Zhao, Ashish Rao, Atri Rudra, and Christopher R{\'e}.
\newblock Just read twice: Closing the recall gap for recurrent language
  models.
\newblock \emph{arXiv preprint arXiv:2407.05483}, 2024.

\bibitem[Baldrati et~al.(2023)Baldrati, Agnolucci, Bertini, and
  Del~Bimbo]{baldrati2023searle}
Alberto Baldrati, Lorenzo Agnolucci, Marco Bertini, and Alberto Del~Bimbo.
\newblock Zero-shot composed image retrieval with textual inversion.
\newblock In \emph{IEEE/CVF International Conference on Computer Vision}, 2023.

\bibitem[BehnamGhader et~al.(2024)BehnamGhader, Adlakha, Mosbach, Bahdanau,
  Chapados, and Reddy]{behnamghader2024llm2vec}
Parishad BehnamGhader, Vaibhav Adlakha, Marius Mosbach, Dzmitry Bahdanau,
  Nicolas Chapados, and Siva Reddy.
\newblock {LLM2Vec}: Large language models are secretly powerful text encoders.
\newblock In \emph{Conference on Language Modeling}, 2024.

\bibitem[Dai et~al.(2023)Dai, Li, Li, Tiong, Zhao, Wang, Li, Fung, and
  Hoi]{dai2023instructblip}
Wenliang Dai, Junnan Li, Dongxu Li, Anthony Meng~Huat Tiong, Junqi Zhao,
  Weisheng Wang, Boyang Li, Pascale Fung, and Steven Hoi.
\newblock {InstructBLIP}: Towards general-purpose vision-language models with
  instruction tuning.
\newblock \emph{arXiv preprint arXiv:2305.06500}, 2023.

\bibitem[Dehghani et~al.(2019)Dehghani, Gouws, Vinyals, Uszkoreit, and
  Kaiser]{dehghani2019universal}
Mostafa Dehghani, Stephan Gouws, Oriol Vinyals, Jakob Uszkoreit, and {\L}ukasz
  Kaiser.
\newblock Universal transformers.
\newblock In \emph{International Conference on Learning Representations}, 2019.

\bibitem[Ganz et~al.(2024)Ganz, Kittenplon, Aberdam, Ben~Avraham, Nuriel,
  Mazor, and Litman]{ganz2024qavit}
Roy Ganz, Yair Kittenplon, Aviad Aberdam, Elad Ben~Avraham, Oren Nuriel, Shai
  Mazor, and Ron Litman.
\newblock Question aware vision transformer for multimodal reasoning.
\newblock \emph{arXiv preprint arXiv:2402.05472}, 2024.

\bibitem[Geiping et~al.(2025)Geiping, McLeish, Jain, Kirchenbauer, Singh,
  Bartoldson, Kailkhura, Bhatele, and Goldstein]{geiping2025huginn}
Jonas Geiping, Sean McLeish, Neel Jain, John Kirchenbauer, Siddharth Singh,
  Brian~R. Bartoldson, Bhavya Kailkhura, Abhinav Bhatele, and Tom Goldstein.
\newblock Scaling up test-time compute with latent reasoning: A recurrent depth
  approach.
\newblock \emph{arXiv preprint arXiv:2502.05171}, 2025.

\bibitem[Hu et~al.(2021)Hu, Shen, Wallis, Allen-Zhu, Li, Wang, Wang, and
  Chen]{hu2021lora}
Edward~J. Hu, Yelong Shen, Phillip Wallis, Zeyuan Allen-Zhu, Yuanzhi Li, Shean
  Wang, Lu Wang, and Weizhu Chen.
\newblock {LoRA}: Low-rank adaptation of large language models.
\newblock \emph{arXiv preprint arXiv:2106.09685}, 2021.

\bibitem[Huang et~al.(2026{\natexlab{a}})Huang, Shi, Chen, Wen, Liu, Xing, and
  Ma]{huang2026loopedI}
Benhao Huang, Chufan Shi, Junlin Chen, Shicheng Wen, Zhengzhong Liu, Eric Xing,
  and Xuezhe Ma.
\newblock Towards looped models done right, part {I}: Topology, input
  injection, recurrent-state design.
\newblock \emph{arXiv preprint arXiv:2610.23036}, 2026{\natexlab{a}}.

\bibitem[Huang et~al.(2026{\natexlab{b}})Huang, Shi, Chen, Wen, Liu, Xing, and
  Ma]{huang2026loopedII}
Benhao Huang, Chufan Shi, Junlin Chen, Shicheng Wen, Zhengzhong Liu, Eric Xing,
  and Xuezhe Ma.
\newblock Towards looped models done right, part {II}: Rethinking at fixed
  points.
\newblock \emph{arXiv preprint arXiv:2610.06833}, 2026{\natexlab{b}}.

\bibitem[Jiang et~al.(2024)Jiang, Meng, Yang, Yavuz, Zhou, and
  Chen]{jiang2024vlm2vec}
Ziyan Jiang, Rui Meng, Xinyi Yang, Semih Yavuz, Yingbo Zhou, and Wenhu Chen.
\newblock {VLM2Vec}: Training vision-language models for massive multimodal
  embedding tasks.
\newblock \emph{arXiv preprint arXiv:2410.05160}, 2024.

\bibitem[Lan et~al.(2020)Lan, Chen, Goodman, Gimpel, Sharma, and
  Soricut]{lan2020albert}
Zhenzhong Lan, Mingda Chen, Sebastian Goodman, Kevin Gimpel, Piyush Sharma, and
  Radu Soricut.
\newblock {ALBERT}: A lite {BERT} for self-supervised learning of language
  representations.
\newblock In \emph{International Conference on Learning Representations}, 2020.

\bibitem[Li et~al.(2026)Li, Zhang, Long, Chen, Song, Bai, Yang, Xie, Yang, Liu,
  Zhou, and Lin]{li2026qwen3vlembedding}
Mingxin Li, Yanzhao Zhang, Dingkun Long, Keqin Chen, Sibo Song, Shuai Bai,
  Zhibo Yang, Pengjun Xie, An Yang, Dayiheng Liu, Jingren Zhou, and Junyang
  Lin.
\newblock {Qwen3-VL-Embedding} and {Qwen3-VL-Reranker}: A unified framework for
  state-of-the-art multimodal retrieval and ranking.
\newblock \emph{arXiv preprint arXiv:2601.04720}, 2026.

\bibitem[Lin et~al.(2014)Lin, Maire, Belongie, Hays, Perona, Ramanan,
  Doll{\'a}r, and Zitnick]{lin2014coco}
Tsung-Yi Lin, Michael Maire, Serge Belongie, James Hays, Pietro Perona, Deva
  Ramanan, Piotr Doll{\'a}r, and C.~Lawrence Zitnick.
\newblock Microsoft {COCO}: Common objects in context.
\newblock In \emph{European Conference on Computer Vision}, 2014.

\bibitem[Liu et~al.(2023{\natexlab{a}})Liu, Li, Wu, and Lee]{liu2023llava}
Haotian Liu, Chunyuan Li, Qingyang Wu, and Yong~Jae Lee.
\newblock Visual instruction tuning.
\newblock In \emph{Advances in Neural Information Processing Systems},
  2023{\natexlab{a}}.

\bibitem[Liu et~al.(2023{\natexlab{b}})Liu, Feng, Fu, Chen, and
  Wang]{liu2023edis}
Siqi Liu, Weixi Feng, Tsu-jui Fu, Wenhu Chen, and William~Yang Wang.
\newblock {EDIS}: Entity-driven image search over multimodal web content.
\newblock \emph{arXiv preprint arXiv:2305.13631}, 2023{\natexlab{b}}.

\bibitem[Liu et~al.(2021)Liu, Rodriguez-Opazo, Teney, and Gould]{liu2021cirr}
Zheyuan Liu, Cristian Rodriguez-Opazo, Damien Teney, and Stephen Gould.
\newblock Image retrieval on real-life images with pre-trained
  vision-and-language models.
\newblock In \emph{IEEE/CVF International Conference on Computer Vision}, 2021.

\bibitem[McLeish et~al.(2025)McLeish, Li, Kirchenbauer, Kalra, Bartoldson,
  Kailkhura, Schwarzschild, Geiping, Goldstein, and
  Goldblum]{mcleish2025retrofitted}
Sean McLeish, Ang Li, John Kirchenbauer, Dayal~Singh Kalra, Brian~R.
  Bartoldson, Bhavya Kailkhura, Avi Schwarzschild, Jonas Geiping, Tom
  Goldstein, and Micah Goldblum.
\newblock Teaching pretrained language models to think deeper with retrofitted
  recurrence.
\newblock \emph{arXiv preprint arXiv:2511.07384}, 2025.

\bibitem[Meng et~al.(2025)Meng, Jiang, Liu, Su, Yang, Fu, Qin, Chen, Xu, Xiong,
  Zhou, Chen, and Yavuz]{meng2025vlm2vecv2}
Rui Meng, Ziyan Jiang, Ye Liu, Mingyi Su, Xinyi Yang, Yuepeng Fu, Can Qin,
  Zeyuan Chen, Ran Xu, Caiming Xiong, Yingbo Zhou, Wenhu Chen, and Semih Yavuz.
\newblock {VLM2Vec-V2}: Advancing multimodal embedding for videos, images, and
  visual documents.
\newblock \emph{arXiv preprint arXiv:2507.04590}, 2025.

\bibitem[Qian et~al.(2026)Qian, Gong, Xu, Li, Wang, Liu, Qian, Pei, Xu, Luo,
  Wang, Yang, Liu, Ma, Yan, and Han]{qian2026loopvl}
Zhe Qian, Ziyang Gong, Zhongxing Xu, Hehan Li, Zhonghua Wang, Hanmeng Liu,
  Liwei Qian, Xin Pei, Peizhi Xu, Fei Luo, Mingxuan Wang, Xue Yang, Shiwei Liu,
  Yanbiao Ma, Junchi Yan, and Jungong Han.
\newblock {LoopVL}: Recurrent visual intelligence.
\newblock \emph{arXiv preprint arXiv:2609.38426}, 2026.

\bibitem[{Qwen Team}(2026)]{qwen2026qwen35}
{Qwen Team}.
\newblock {Qwen3.5}: Towards native multimodal agents.
\newblock \url{https://github.com/QwenLM/Qwen3.5}, 2026.
\newblock Official open model release.

\bibitem[Schwenk et~al.(2022)Schwenk, Khandelwal, Clark, Marino, and
  Mottaghi]{schwenk2022aokvqa}
Dustin Schwenk, Apoorv Khandelwal, Christopher Clark, Kenneth Marino, and
  Roozbeh Mottaghi.
\newblock {A-OKVQA}: A benchmark for visual question answering using world
  knowledge.
\newblock \emph{arXiv preprint arXiv:2206.01718}, 2022.

\bibitem[Song et~al.(2026)Song, Li, Zhang, Long, Xie, Nie, Zhao, and
  Wu]{song2026uembed}
Tingyu Song, Mingxin Li, Yanzhao Zhang, Dingkun Long, Pengjun Xie, Zhijie Nie,
  Yilun Zhao, and Shu Wu.
\newblock {UEmbed}: Unified sparse and dense multimodal embeddings.
\newblock \emph{arXiv preprint arXiv:2608.02583}, 2026.

\bibitem[Springer et~al.(2025)Springer, Kotha, Fried, Neubig, and
  Raghunathan]{springer2024echo}
Jacob~Mitchell Springer, Suhas Kotha, Daniel Fried, Graham Neubig, and Aditi
  Raghunathan.
\newblock Repetition improves language model embeddings.
\newblock In \emph{International Conference on Learning Representations}, 2025.

\bibitem[Wang et~al.(2024)Wang, Bai, Tan, Wang, Fan, Bai, Chen, Liu, Wang, Ge,
  Fan, Dang, Du, Ren, Men, Liu, Zhou, Zhou, and Lin]{wang2024qwen2vl}
Peng Wang, Shuai Bai, Sinan Tan, Shijie Wang, Zhihao Fan, Jinze Bai, Keqin
  Chen, Xuejing Liu, Jialin Wang, Wenbin Ge, Yang Fan, Kai Dang, Mengfei Du,
  Xuancheng Ren, Rui Men, Dayiheng Liu, Chang Zhou, Jingren Zhou, and Junyang
  Lin.
\newblock {Qwen2-VL}: Enhancing vision-language model's perception of the world
  at any resolution.
\newblock \emph{arXiv preprint arXiv:2409.12191}, 2024.

\bibitem[Wang et~al.(2026)Wang, Hao, Li, Cao, Tang, and Guo]{wang2026reloop}
Shijie Wang, Xiangzhao Hao, Yueti Li, Guangyu Cao, Xinyu Tang, and Haiyun Guo.
\newblock {ReLoop-UME}: Recurrent depth with learnable retrieval registers for
  universal multimodal embedding.
\newblock \emph{arXiv preprint arXiv:2607.28751}, 2026.

\bibitem[Wu et~al.(2021)Wu, Gao, Guo, Al-Halah, Rennie, Grauman, and
  Feris]{wu2021fashioniq}
Hui Wu, Yupeng Gao, Xiaoxiao Guo, Ziad Al-Halah, Steven Rennie, Kristen
  Grauman, and Rogerio Feris.
\newblock Fashion {IQ}: A new dataset towards retrieving images by natural
  language feedback.
\newblock In \emph{IEEE/CVF Conference on Computer Vision and Pattern
  Recognition}, 2021.

\bibitem[Yang et~al.(2025)Yang, Kautz, and Hatamizadeh]{yang2024gateddeltanet}
Songlin Yang, Jan Kautz, and Ali Hatamizadeh.
\newblock Gated delta networks: Improving {Mamba2} with delta rule.
\newblock In \emph{International Conference on Learning Representations}, 2025.

\bibitem[Zhou et~al.(2026)Zhou, Mei, Li, Wang, Rao, and Lyu]{zhou2026wemm}
Junjie Zhou, Ke Mei, Lei Li, Tianyi Wang, Fengyun Rao, and Jing Lyu.
\newblock {WeMM-Embedding}: {WeChat} multi-modal embedding technical report.
\newblock \emph{arXiv preprint arXiv:2608.24053}, 2026.

\bibitem[Zhu et~al.(2025)Zhu, Wang, Hua, Zhang, Li, Que, Wei, Wen, Yin, Xing,
  Li, Shi, Ma, Li, Kergan, Smith, Qu, Hui, Wu, Min, Huang, Zhou, Ye, Liu, Yang,
  Shi, Lin, Zhao, Cai, Zhang, Huang, Bengio, and Eshraghian]{zhu2025ouro}
Rui-Jie Zhu, Zixuan Wang, Kai Hua, Tianyu Zhang, Ziniu Li, Haoran Que, Boyi
  Wei, Zixin Wen, Fan Yin, He Xing, Lu Li, Jiajun Shi, Kaijing Ma, Shanda Li,
  Taylor Kergan, Andrew Smith, Xingwei Qu, Mude Hui, Bohong Wu, Qiyang Min,
  Hongzhi Huang, Xun Zhou, Wei Ye, Jiaheng Liu, Jian Yang, Yunfeng Shi,
  Chenghua Lin, Enduo Zhao, Tianle Cai, Ge Zhang, Wenhao Huang, Yoshua Bengio,
  and Jason Eshraghian.
\newblock Scaling latent reasoning via looped language models.
\newblock \emph{arXiv preprint arXiv:2510.25741}, 2025.

\bibitem[Zhu et~al.(2016)Zhu, Groth, Bernstein, and Fei-Fei]{zhu2016visual7w}
Yuke Zhu, Oliver Groth, Michael Bernstein, and Li Fei-Fei.
\newblock {Visual7W}: Grounded question answering in images.
\newblock In \emph{IEEE Conference on Computer Vision and Pattern Recognition},
  2016.

\end{thebibliography}
}
\end{document}